\def\year{2022}\relax
\documentclass[letterpaper]{article} 
\usepackage{aaai22}  
\usepackage{times}  
\usepackage{helvet}  
\usepackage{courier}  
\usepackage[hyphens]{url}  
\usepackage{graphicx} 
\usepackage{natbib}  
\usepackage{caption} 
\DeclareCaptionStyle{ruled}{labelfont=normalfont,labelsep=colon,strut=off} 
\usepackage{algorithm}
\usepackage{algorithmic}

\usepackage{newfloat}
\usepackage{listings}
\floatstyle{ruled}
\newfloat{listing}{tb}{lst}{}
\floatname{listing}{Listing}
\nocopyright
\usepackage{booktabs}
\usepackage{amsmath}
\usepackage{amsfonts}
\usepackage[inline]{enumitem} 
\usepackage{multirow} 

\title{Invertible Frowns: Video-to-Video Facial Emotion Translation}
\author {
    Ian Magnusson,\textsuperscript{\rm 1}\equalcontrib
    Aruna Sankaranarayanan, \textsuperscript{\rm 2}\equalcontrib
    Andrew Lippman \textsuperscript{\rm 2}
}
\affiliations {
    \textsuperscript{\rm 1} Northeastern University, Boston, Massachusetts, USA\\
    \textsuperscript{\rm 2} Media Lab (MIT), Cambridge, Massachusetts, USA\\
    magnusson.i@northeastern.edu, arunas@mit.edu, lip@media.mit.edu
}

\begin{document}
\maketitle

\begin{abstract}
We present Wav2Lip-Emotion, a video-to-video translation architecture that modifies facial expressions of emotion in videos of speakers. Previous work modifies emotion in images, uses a single image to produce a video with animated emotion, or puppets facial expressions in videos with landmarks from a reference video. However, many use cases such as modifying an actor's performance in post-production, coaching individuals to be more animated speakers, or touching up emotion in a teleconference require a video-to-video translation approach. We explore a method to maintain speakers' lip movements, identity, and pose while translating their expressed emotion. Our approach extends an existing multi-modal lip synchronization architecture to modify the speaker's emotion using L1 reconstruction and pre-trained emotion objectives. We also propose a novel automated emotion evaluation approach and corroborate it with a user study. These find that we succeed in modifying emotion while maintaining lip synchronization. Visual quality is somewhat diminished, with a trade off between greater emotion modification and visual quality between model variants. Nevertheless, we demonstrate \begin{enumerate*}[label=(\arabic*)] \item that facial expressions of emotion can be modified with nothing other than L1 reconstruction and pre-trained emotion objectives and \item that our automated emotion evaluation approach aligns with human judgements. \end{enumerate*}
\end{abstract}

\section{Introduction}

\begin{figure}[!htb]
\begin{center}
\includegraphics[width=1\linewidth]{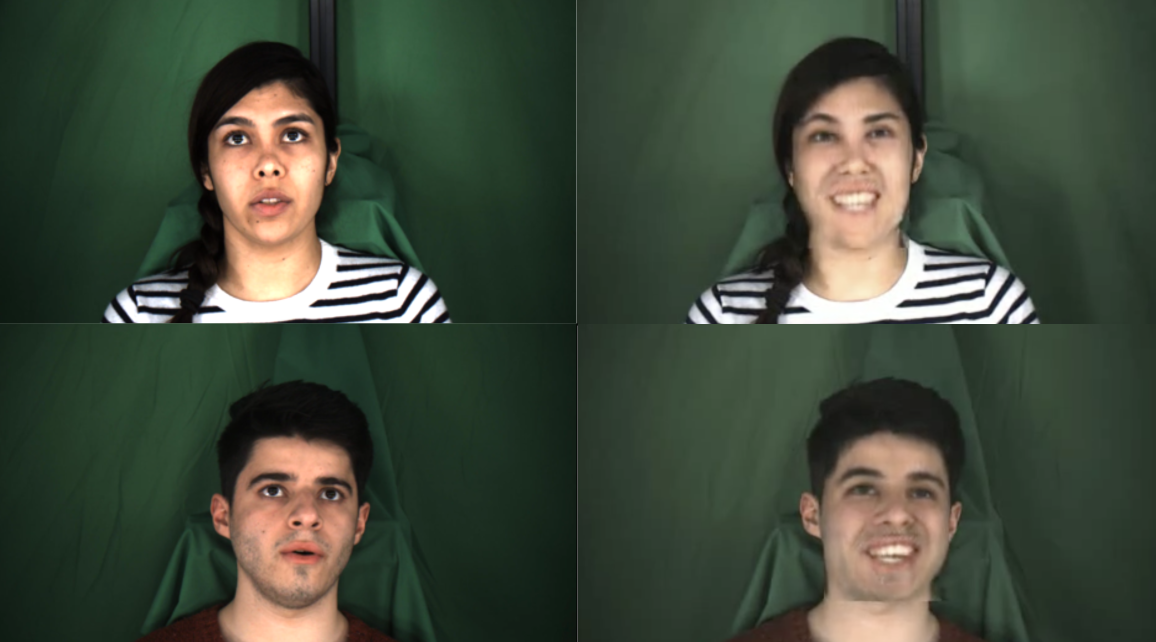}
\end{center}
\caption{Wav2Lip-Emotion takes existing videos (real neutral inputs on left) and modifies facial expressions of emotion (generated happy outputs on right) while maintaining speakers' lip synchronization and pose. Examples here are from all-around best model on unseen faces.}
\label{fig:intro_example}
\end{figure}

The face is the window to the mind: the human face conveys information about our mental, physical, and of most of all, the emotional state. We can sense if someone is in emotional distress or happily animated. Using facial expression, voice, muscular tension, and other cues, we are also able to pick up on subtler emotions. Research has shown that our facial expressions can invoke an emotion in both expresser \cite{izard_facial_nodate, ekman1993facial} and the recipient of the expression \cite{brunson2002impact}. Further, research on non-verbal behaviours has shown that facial displays are integral to how we signal and maintain social dominance \cite{keating1977facial} and other social cues. Even when we are tasked with maintaining a neutral facial disposition (e.g., newscasters), we often display subtle cues that convey our biases \cite{miller_election_facial}.

In several scenarios, the expression of facial emotion may be suppressed. For example, in studies inhibited children have lower facial expressiveness than uninhibited children \cite{kagan1993temperamental}. Their baseline facial dynamics operate over a smaller range \cite{picard2000affective}. Facial emotion suppression is also inhibited by post-traumatic stress disorder (PTSD), as measured by the movements of certain facial muscles during a facial emotion recognition task \cite{passardi_facial_2019}. Facial palsy interferes with expressing facial emotions as well; computer vision models that are trained to recognize emotions can often perceive less joy and greater negative emotions in the faces of facial palsy patients \cite{dusseldorp2019eye}.

Thus, the artificial suppression or augmentation of facial emotion may be desirable at times. Individuals with or without inhibited facial expressions may benefit from tuning their own expressions to better fit their social circumstances. One may want to alter the expressions in videos shown to them. Speakers might be yelling at each other during a video conference, but nevertheless want to gather the content in their exchange without the unpleasant expressions. Or a film director may want to augment or diminish the expressions of an actor.

Inspired by this idea, our primary contribution is the creation of a deep learning model that modifies the facial emotion of a speaker in a given video, while preserving the visemes, pose, and identity from the original video. Existing research in this area has produced models that modify emotion in an image,
synthesize videos with a certain emotion from a single image,
or puppet expressions using reference input.
However, to our knowledge no work exists that directly modifies the emotion in an existing video.

We build Wav2Lip-Emotion,
\footnote{Available at https://github.com/jagnusson/Wav2Lip-Emotion.} 
a multi-modal computer vision model that modifies facial emotion via video-to-video translation. This model adds emotion modification to Wav2Lip \cite{prajwal2020wav2lip}, a recent model that synchronizes face videos with speech audio by means of L1 reconstruction and pre-trained synchronization objectives. We also propose an automated emotion evaluation using the NISL2020 model \cite{Deng2020NISL2020} to examine continuous valued changes in valence and arousal in generated videos against baseline changes in pairs of emotions performed by human actors in the MEAD dataset \cite{kaisiyuan2020mead}. Through the automated evaluation backed up by a user study,  we demonstrate that facial emotion can be modified while maintaining lip sync and moderate visual quality by L1 reconstruction and pre-trained emotion objectives alone.

\section{Related Work}
\paragraph{Image-to-image translation}
One approach utilizes the StyleGAN architecture \cite{Karras_2019_CVPR} to modify emotion in an image by first using optimization to recover a latent StyleGAN vector that approximates the input image and then shifting it along a learned subspace for the desired emotion shift \cite{NikitkoStyleGANencoder}. Other researchers have employed 3D face synthesis techniques to generate various emotions on a neutral face \cite{Chen_2019_ICCV} or used cycle consistency loss to overcome the availability of paired emotion data \cite{pumarolaGANimation2018}.

\paragraph{Image-to-video translation} Synthesizing a video from a single image is another context in which emotion manipulation has been explored. Fan et al.  \cite{Fan_Huang_Gan_Huang_Gong_2019} utilize a dataset of short videos of faces moving from a neutral expression to a specified emotion expression to train a controlled image-to-image translator. The translator reconstructs the expression video frame by frame using the neutral expression in the first frame and the frame index offset of the frame to be generated. Most similarly to our work, the authors of the MEAD dataset \cite{kaisiyuan2020mead} also make a baseline model that generates a "talking head" video using audio and a face image. One module that generates lip movements in the lower face based on the audio, while another modifies emotion in the upper face. Both halves are combined using a refinement network. Such talking head generation is suitable for cases such as animating memes where a full speaker video is not available.

\paragraph{Video-to-video translation} 
To our knowledge no work in video-to-video translation directly tackles the problem of emotion manipulation. Other video-to-video translation work maps the pose and facial keypoints of an input driving video onto a different identity provided by a single image or video  \cite{NEURIPS2019_31c0b36a}. This enables the puppeting of emotional expressions and lip movements on the basis of a separate video. One clever work uses a large repository of pre-annotated reference videos to quickly look up instances of relevant phonemes with desired facial expressions which can then be used with such a puppeting technique to edit an input video \cite{yao2021talkinghead}. While highly efficient, this approach is limited to repeating already recorded expressions. Instead we adapt the approach of Wav2Lip which utilizes an L1 reconstruction loss and a pre-trained discriminator derived from SyncNet \cite{Chung2016SyncNet} to translate out of sync lip videos to synchronized ones \cite{prajwal2020wav2lip}. We extend their model to emotion modification.

\section{Approach}
Our Wav2Lip-Emotion approach extends Wav2Lip \cite{prajwal2020wav2lip} to modify emotion via L1 reconstruction and pre-trained emotion objectives, while maintaining lip synchronization, pose, and identity.

\begin{enumerate}
    \item We modify the Wav2Lip architecture to use a pre-trained emotion classifier as an additional objective. We train to maximize the class likelihood of the desired emotion. 
    
    \item We fine-tune Wav2Lip-Emotion using videos from the MEAD controlled emotion dataset \cite{kaisiyuan2020mead}. We take input videos from a specified \textit{source emotion} and we sample the target videos from a different \textit{destination emotion} to allow the L1 reconstruction loss to also encourage emotion modification. For the scope of this work we only produce models that modify emotion between a pair of specific emotions.
    
    \item The Wav2Lip architecture masks the lower half of target image frames to create a pose prior input that does not reveal lip information. In Wav2Lip-Emotion, we experiment with masking the entire face to conceal target emotion beyond the mouth. Likewise, we modify Wav2Lip's visual quality discriminator to scrutinize the whole rather than bottom half of the frame.
    
    \item At inference, unlike the original Wav2Lip architecture which synchronizes the lip movements in a video to unsynchronized audio, Wav2Lip-Emotion retains the original audio and simply modifies the emotion while ensuring lip synchronization.
    
\end{enumerate}

\subsection{Datasets}


The Wav2Lip architecture is trained on the LRS2 dataset \cite{LRS2} which contains thousands of spoken sentences from BBC television. To introduce emotion modification, we fine tune the Wav2Lip architecture on MEAD, a dataset of actors performing a set of utterances with several emotional variations that span arousal and valence.

The MEAD dataset 
is a controlled emotion dataset consisting of 40 hours of videos of 60 actors reading the same sentence with different facial expressions. The dataset is extensive -- it includes a diverse set of actors from different continents spanning 15 different countries. The speakers therefore have several regional characteristics in their faces and their manner of speaking.

As of May 2021, only the first part of the dataset containing videos of 47 individuals has been released. Each individual performs the following emotions at 3 levels of intensity: angry, disgust, fear, contempt, happy, sad, and surprise. Neutral is used as an eighth emotion but only has one intensity level. The recognizability of the emotions is validated by a user study that finds that labellers can accurately identify the intended emotion performed by the actor. All videos are framed around the head, from identical angles for all actors, with controlled lighting.

In this paper, we describe results achieved from training our model with only the happy and sad level 3 and neutral level 1 emotions. We limit our analysis to these 3 emotions with 1 intensity level each and only frontal videos due to the large amount of compute required to pre-process the data. Happy and sad emotions have highest levels of recognizability, while neutral emotion on the other hand was much lower. We include these emotions to capture a selection of easier and harder emotion modifications in our analysis.



\subsection{Model}

Our Wav2Lip-Emotion extends the Wav2Lip architecture with an additional emotion objective and modifies the visual quality discriminator. Thus the model is composed of a generator and 3 discriminators: \begin{enumerate*}[label=(\arabic*)]
    \item lip synchronization and
    \item emotion pre-trained objectives as well as
    \item an adversarially trained visual quality objective.
\end{enumerate*}
The model operates on inputs composed of short windows of audio and face-cropped video frames, and outputs generated face frames.

\subsubsection{Generator}
The generator, $G$ contains 3 blocks. An identity encoder, speech encoder, and face decoder. The architectural details for these blocks are outlined in the Wav2Lip paper. The identity encoder concatenates a random frame with a pose prior consisting of a masked version of the target face. The speech signal is also encoded as a stack of 2D convolutions which are concatenated with the frame. The decoder is also a set of convolutional layers that have been modified for upsampling. 

\subsubsection{Original Training Objectives}
Wav2Lip follows other image translation work in using the L1 reconstruction loss between the real ($\mathbf{F_G}^i$) and generated frames ($\mathbf{F_g}^i$):
\begin{align}
L_{r} = \frac {1}{N} \sum_{i=1}^{N} \left \|\mathbf{F_g}^i - \mathbf{F_G}^i\right \|_1
\end{align}

The pre-trained lip synchronization discriminator used by the authors of Wav2Lip is a modified version of SyncNet \cite{Chung2016SyncNet}, a model that detects lip synchronization errors in videos, which is pre-trained on the LRS2 dataset. Note that this lip synchronization discriminator is not trained further during the Wav2Lip or Wav2Lip-Emotion training process.
Its loss function uses the cosine similarity between the speech audio ($\mathbf{s}^i$) and the face video ($\mathbf{v}^i$) with binary cross-entropy loss:
\begin{align}
P_{s}^i = \frac {\mathbf{v}^i \cdot \mathbf{s}^i}{max(\left \|\mathbf{v}^i\right \|_2 \cdot \left \|\mathbf{s}^i\right \|_2, \epsilon )}
\end{align}
\begin{align}
E_{s} = \frac {1}{N} \sum_{i=1}^{N} - log(P^{i}_{s})
\end{align}

Wav2Lip also uses a visual quality discriminator, which penalizes unrealistic faces. Unlike the pre-trained lip synchronization discriminator, the visual quality discriminator learns by discriminating the generated images and real images during training.
The discriminator is trained to maximize the $L_{d}$ objective function.
\begin{align}
L_{g} = \mathbb{E}_{\mathbf{x} \sim \mathbf{F}_g} [log(1-D(\mathbf{x}))]
\end{align}
\begin{align}
L_{d} = \mathbb{E}_{\mathbf{x} \sim \mathbf{F}_G} [log(D(\mathbf{x}))] + L_{g}
\end{align}

Originally the visual quality discriminator only examines the lower half of the face. But since emotion modification also takes place in the upper half, we modify the architecture to discriminate the whole image. We also introduce residual connections to overcome problems with vanishing gradients that occurred during the process of retraining the model on LRS2 data with this modified discriminator architecture.

\subsubsection{Pre-trained Emotion Objective}
In addition to the two discriminators in the original model, we add a pre-trained emotion objective. We utilize a DenseNet model \cite{huang2017densenet} trained by \cite{luanresmaskingnet2020}. This emotion classifier detects 6 emotions and neutral expressions and achieves accuracy of 73.16\% on the FER2013 dataset \cite{giannopoulos2018deep}. Due to pre-processing compute constraints, we utilize only happy, neutral, and sad. We put the logits, $\mathbf{z}$, output by the final layer through softmax to form the likelihood of each emotion class $e \in E$:
\begin{align}
g_e = {\exp(z_e) \over \sum_{k \in E }\exp(z_k)}
\end{align}
    
We minimize the deviation of the desired emotion class likelihood, $g_d$ from the maximum value:

\begin{align}
    L_{e} = \frac {1}{N} \sum_{i=1}^{N} 1 - g_d^i
\end{align}    

\subsubsection{Total Loss}
The overall loss minimized by the generator is given by the weighted sum, 
\begin{align}
L_{total} = s_r \cdot L_{r} + s_w \cdot E_{s} + s_g \cdot L_{g} + s_e \cdot L_{e}
\end{align}
where $s_r$, $s_w$, $s_g$, and $s_e$ are weight hyperparameters.

\subsection{Variants}

\subsubsection{Masking} We present two model variants with respect to our masking strategy. These are the \textbf{full} masking approach which covers the full face in the input pose prior to conceal all emotion information, and \textbf{half} masking which preserves the original Wav2Lip masking strategy and enables the use of pre-trained Wav2Lip checkpoints.
To create full masks, we use the \textit{dlib} library \cite{dlib09} along with a landmark detector \cite{codeniko2019landmarks} that identifies a set of 81 facial keypoints.
We then mask a convex hull along the boundary of the face using the list of facial boundary keypoints. 

\subsubsection{Emotion Modification Strategy} Our work explores two avenues for encouraging emotion modification in generated videos: L1 reconstruction loss and a pre-trained emotion objective. Thus we explore 3 variants—\textbf{L1}, \textbf{Emotion Objective}, and \textbf{L1 + Emotion Objective}. The impact of the emotion objective can be ablated by simply removing that objective. Since the L1 reconstruction loss is also critical for maintaining lip synchronization and visual quality, we cannot simply remove it. It is able to encourage emotion modification only when we use the MEAD emotion annotations to provide target video frames drawn from a specified \textit{destination emotion} which differs from a \textit{source emotion} that is used for the input reference image. So to ablate the emotion modification via L1 loss we use \textit{source emotion} video frames for both inputs and targets.

\begin{figure*}[!htb]
\centering
\begin{minipage}{.28\textwidth} 
\includegraphics[width=1\linewidth]{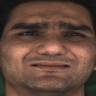}
\end{minipage}%
\begin{minipage}{.42\textwidth} 
\includegraphics[width=1\linewidth]{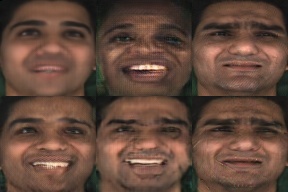}
\end{minipage}%
\begin{minipage}{.28\textwidth}
\includegraphics[width=1\linewidth]{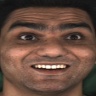}
\end{minipage}
\caption{Input sad frame (left), generated happy outputs (grid center), and target ground truth happy frame (right). In the grid the first row is \textbf{Full} masking while the second row is \textbf{Half} masking. The columns are emotion modification strategies left to right as follows: \textbf{L1 + Emotion Objective}, \textbf{L1}, \textbf{Emotion Objective}.}

\label{fig:test_examples}
\end{figure*}

\section{Evaluation}

We propose an automated emotion evaluation approach that compares changes in valence and arousal in generated videos against changes in baseline human performances in MEAD. We also provide qualitative examples (see Figure \ref{fig:test_examples}) and a user study with findings that corroborate our automated emotion evaluation approach. We also provide example videos in the supplementary material.

\subsection{Wav2Lip Retraining}
\label{Wav2Lip Retraining}

\begin{table}[htb]
\centering
\begin{tabular}{c ccc }
\toprule
\textbf{masking} & \textbf{LSE-D$\downarrow$} & \textbf{LSE-C$\uparrow$} &\textbf{FID$\downarrow$}\\ \midrule
full & 7.640 & 6.213 &  5.938 \\
half & 7.013 & 7.029 &  7.818 \\
\midrule
\textbf{original model} &&& \\
\midrule
Wav2Lip + GAN & \textbf{6.469} & \textbf{7.781} & \textbf{4.446} \\
\bottomrule

\end{tabular}
\caption{Lip sync (\textbf{LSE-D}, \textbf{LSE-C}) and visual quality (\textbf{FID}) for re-trained Wav2Lip with modified visual quality discriminator necessary for emotion modification, reported on LRS2 dataset compared with original Wav2Lip results.}
\label{tab:lrs2_retrain_results}
\end{table}
\begin{table*}[htb]
\centering
\begin{tabular}{cc cc c cc }
\toprule
\textbf{masking} & \textbf{emotion modification strategy} & \textbf{LSE-D$\downarrow$} & \textbf{LSE-C$\uparrow$} &\textbf{FID$\downarrow$} & \textbf{$\Delta$ valence $\uparrow$} & \textbf{$\Delta$ arousal $\uparrow$}   \\ \midrule
full&L1 + Emotion Objective&11.357&1.378&103.016&0.729&0.400 \\
full&Emotion Objective&10.993&1.855&73.046&0.044&0.200 \\
full&L1&11.413&1.477&141.719&0.995&\textbf{0.803} \\
half&L1 + Emotion Objective&10.563&2.101&71.508&\textbf{1.000}&0.522 \\
half&Emotion Objective&\textbf{10.443}&\textbf{2.328}&\textbf{48.819}&0.095&0.244 \\
half&L1&10.673&2.000&82.709&0.945&0.595 \\
\midrule
\multicolumn{2}{c}{\textbf{Average Wav2Lip Original}}& 10.032 & 2.751&	31.813& - & -  \\
\multicolumn{2}{c}{\textbf{Average Ground Truth}}& 10.651 &2.533&	26.302& -  & -  \\
\bottomrule

\end{tabular}
\caption{Comparison averaged across emotions of Wav2Lip-Emotion variants and ground truth on lip sync (\textbf{LSE-D}, \textbf{LSE-C}), visual quality (\textbf{FID}), and our novel automated emotion evaluation approach that normalizes change in valence and arousal by a human baseline. Original model and ground truth shown for contrast, with FID averaged over emotion combinations.}
\label{tab:top_level_results}
\end{table*}
Since we redesign the architecture of the visual quality discriminator and introduce a new full face masking strategy in one of our variants, it was necessary to retrain Wav2Lip on the LRS2 data used by the original authors. For our \textbf{full} masking variants, we retrained Wav2Lip from scratch following the same procedure as the original authors except for changing the masking technique and the visual quality discriminator. For our \textbf{half} masking variants, we were able to start training from checkpoints provided by the Wav2Lip authors before the introduction of the visual quality discriminator objective. 

In Table \ref{tab:lrs2_retrain_results} we report the results of retraining Wav2Lip on LRS2 prior to fine-tuning for emotion modification. We give the three metrics employed by the original authors: lip synchronization error distance (\textbf{LSE-D}), lip synchronization error confidence (\textbf{LSE-C}), and Fréchet Inception distance (\textbf{FID}).

LSE-D and LSE-C utilize SyncNet \cite{Chung2016SyncNet}, the original architecture modified by Wav2Lip for their lip synchronization discriminator. The originally reported accuracy of this model is over 99\%. The LSE-D measures the L2 distance between the SyncNet encodings of audio and video, where close vectors are trained to represent synchronization. LSE-C on the other hand represents the SyncNet confidence output, where a larger number indicates better synchronization. 

FID \cite{heusel2018gans} is a commonly used automatic measure of distance between image datasets. Feature representations of the images in the dataset are encoded by the pre-trained Inception network and then the Fréchet distance is calculated between two Gaussians fitted to these representations. Wav2Lip takes the FID of generated output frames against ground truth frames as a measure of overall visual quality, where a lower value indicates higher visual quality from greater fidelity to the distribution of ground truth frames.

While our retrained Wav2Lip models do not achieve identical performance with the original, they are nevertheless sufficiently comparable given that our objective is simply to maintain lip synchronization rather than alter it. We trained all models until convergence.


\subsection{Qualitative Observations}

Figure \ref{fig:test_examples} provides examples of generated frames from all of our model variants on sad to happy emotion modification based on inputs from our test split of the MEAD data. The input sad image frame is shown on the left, the happy generated outputs are provided for each of the 6 model variations in the grid in the center, and a frame from the target ground truth happy performance is provided on the right for contrast. In the grid the top row is \textbf{Full} masking while the bottom row is \textbf{Half} masking. The columns are emotion modification strategies left to right as follows: \textbf{L1 + Emotion Objective}, \textbf{L1}, \textbf{Emotion Objective}.

Analysis of a selection of examples seems to indicate that the \textbf{L1 + Emotion Objective} emotion modification strategy has issues with blurry outputs but performs the task most faithfully by modifying the emotion while preserving pose. The \textbf{L1} emotion modification strategy on the other hand produces more crisp results with clear emotion modification but does a poor job of preserving pose and visual quality. Finally, \textbf{Emotion Objective} emotion modification does an excellent job of preserving pose and sync, but produces very little emotion modification.

\begin{table*}[htb]
\centering
\begin{tabular}{ c ccc cc c cc }
\toprule
&\textbf{emotion modification} & \textbf{LSE-D$\downarrow$} & \textbf{LSE-C$\uparrow$} & \textbf{FID$\downarrow$} & $\mathbf{v_d - v_s}$ &$\mathbf{a_d - a_s}$ & \textbf{$\Delta$ valence $\uparrow$} & \textbf{$\Delta$ arousal $\uparrow$}   \\ \midrule
\parbox{3mm}{\multirow{6}{*}{\rotatebox[origin=c]{90}{\textbf{All Variants}}}}
&happy $\rightarrow$ neutral &11.543&1.52&95.127&-0.746&-0.213&0.649&0.767 \\
&happy $\rightarrow$ sad&10.715&1.986&77.768&-0.807&-0.069&\textbf{2.143}&0.121 \\
&neutral $\rightarrow$ happy&\textbf{10.56}&\textbf{2.277}&93.241&0.74&0.212&0.66&\textbf{0.808} \\
&neutral $\rightarrow$ sad&10.774&2.113&98.442&-0.077&0.155&0.185&0.244 \\
&sad $\rightarrow$ happy&10.82&1.676&\textbf{65.007}&0.807&0.069&0.663&0.377 \\
&sad $\rightarrow$ neutral&11.03&1.566&91.232&0.084&-0.155&-0.486&0.427 \\
\midrule
\parbox{3mm}{\multirow{6}{*}{\rotatebox[origin=c]{90}{\textbf{Best Variant}}}}
&happy $\rightarrow$ neutral&11.017&1.757&72.889&-0.746&-0.213&0.661&0.826\\
&happy $\rightarrow$ sad&\textbf{10.117}&2.312&77.642&-0.807&-0.069&\textbf{3.537}&0.455\\
&neutral $\rightarrow$ happy&10.462&2.449&67.377&0.746&0.213&1.063&\textbf{1.032}\\
&neutral $\rightarrow$ sad&10.375&\textbf{2.576}&69.902&-0.077&0.155&0.383&0.441\\
&sad $\rightarrow$ happy&10.489&1.838&\textbf{68.397}&0.807&0.069&1.246&-0.444\\
&sad $\rightarrow$ neutral&10.918&1.674&72.839&0.077&-0.155&-0.888&0.825 \\

\bottomrule

\end{tabular}
\caption{Comparison over \textit{source} and \textit{destination} emotions for lip sync (\textbf{LSE-D}, \textbf{LSE-C}), visual quality (\textbf{FID}), and change of \textbf{valence} and \textbf{arousal} in generated outputs normalized by human baseline change between ground truth inputs and targets ($\mathbf{v_d - v_s}$ and $\mathbf{a_d - a_s}$). Top section averaged over Wav2Lip-Emotion variants, and bottom section best all-around variant.}
\label{tab:per_emotion_results}
\end{table*}

\subsection{Automatic Evaluations}

We report automated evaluation results for Wav2Lip-Emotion variants in Table \ref{tab:top_level_results}. We follow Wav2Lip's utilization of pre-trained automated evaluation metrics (\textbf{LSE-D}, \textbf{LSE-C}, and \textbf{FID}, explained in Section \ref{Wav2Lip Retraining})
and craft an approach for automatic emotion evaluation (\textbf{$\Delta$ valence} and \textbf{$\Delta$ arousal}).

To measure valence and arousal in videos we use the NISL2020 model \cite{Deng2020NISL2020}, the winner of FG-2020's Competition in Affective Behavior Analysis in-the-wild (ABAW) \cite{kollias2020analysing}. Trained on the Aff-Wild2 dataset \cite{Kollias2018Aff-Wild2}, this model outputs per-frame valence scores, ranging from [-1,1], that indicate how positive or negative an expression is. It also outputs per-frame arousal scores, ranging from [0,1], that indicate how active or calm an expression is. The model incorporates information from a temporal window of frames for each judgment. We take the average of each value over all frames in a video to get a video level score.

In order to make our emotion metric comparable across different pairs of \textit{source} and \textit{destination} emotions, we devise the following normalization scheme. Our test split of the MEAD dataset contains performed \textit{destination emotion} target videos associated with the \textit{source emotion} input videos that can be used as a baseline for our evaluation. Thus we get average valence and arousal scores over all frames of an actor performing a given utterance for the generated outputs ($v_g$ and $a_g$), \textit{source} ground truth emotion ($v_s$ and $a_s$), and \textit{destination} ground truth emotion ($v_d$ and $a_d$). We then take the ratio of the change in the generated video compared to its \textit{source emotion} input against the change in the \textit{destination emotion} video compared to the \textit{source emotion} video:
\begin{align}
    \Delta ~\text{valence} = {{v_g - v_s} \over {v_d - v_s}}
\end{align}
\begin{align}
    \Delta ~\text{arousal} = {{a_g - a_s} \over {a_d - a_s}}
\end{align}
Thus intuitively positive values indicate change in the correct direction, with a value of $1$ indicating a change in emotion identical to the ground truth change performed by the actors in the MEAD dataset. Meanwhile emotion pairs like sad and neutral will have smaller ground truth changes and will be scaled so as not to be drowned out when aggregated with pairs that have larger ground truth shifts like sad and happy. Scores greater than $1$ are possible. While overshooting the valence change towards happy or sad emotion is desirable, overshooting the valance modification towards neutral is not. Thus, we further take $\Delta ~\text{valence}_{*2n} = 1 - |1 - \Delta ~\text{valence}|$ as our normalized score for neutral \textit{destination} modifications to penalize overshooting.

The numbers presented in Table \ref{tab:top_level_results} are the averaged values over all videos in each condition, further micro-averaged over all 6 emotion pairs of $\{\text{sad}, \text{neutral}, \text{happy}\}$. This reveals that no one model variant excels in all cases. Unsurprisingly the \textbf{half} masking with \textbf{Emotion Objective} modification strategy variant performs best on lip synchronization and image quality metrics as this model is closest to the original Wav2Lip approach. It takes advantage of the half masked model checkpoints provided by the authors and needs to retrain only the visual quality discriminator unlike our \textbf{full} masking models which also retrain the generator. This model also achieves better metrics on the LRS2 as shown in Table \ref{tab:lrs2_retrain_results}. However, this model performs relatively poorly on our automated emotion evaluation.

The best performing models on the emotion evaluation appear to come from the \textbf{L1} and \textbf{L1 + Emotion Objective} emotion modification strategy variants rather the \textbf{Emotion Objective} variants. This indicates that the L1 reconstruction loss along with the novel use of \textit{destination emotion} videos for target frames produces most emotion modification, while only a small amount of modification is achieved with the emotion objective alone. Nevertheless the \textbf{Emotion Objective} only variants do still move emotion (particularly arousal) in the right direction and achieve better synchronization and visual quality metrics. The \textbf{L1 + Emotion Objective} variant that utilizes both approaches, appears to provide a balance between synchronization and visual quality against emotion modification.  

We contrast our model performances on the synchronization and visual quality metrics against those same metrics run on the ground truth data as well as the generated outputs of the original Wav2Lip run on the synchronized ground truth video and audio. \textbf{LSE-D} and \textbf{LSE-C} numbers on the ground truth and Wav2Lip generated happy, neutral, and sad videos are averaged over emotions to produce the numbers at the bottom of Table \ref{tab:top_level_results}. The \textbf{FID} scores meanwhile are the average over the FID between all combinations of the three emotions. While our models' synchronization and visual quality performance are worse than that of Wav2Lip on the LRS2 (see Table \ref{tab:lrs2_retrain_results}), our metrics are better aligned for Wav2Lip's outputs on MEAD as well as the metrics on the unmodified ground truth itself. This discrepancy may arise from slight audio desynchronization in the raw MEAD data. Meanwhile the use of FID to judge visual quality may be better suited for lip synchronization than emotion modification, since the latter makes much more salient changes to the images in the videos. Evidently even real differences in performed emotion made by the same speaker in the MEAD ground truth data shift the distributions of Inception encodings more than Wav2Lip's generated results on LRS2. The FID between original Wav2Lip outputs on MEAD and ground truth frames from a distinct emotion is somewhat higher than the FID between ground truth emotions only, suggesting that the FID nevertheless measures some additional differences in visual quality produced by video translation.

\begin{table*}[htb]
\centering
\begin{tabular}{ccccc}
\toprule

\textbf{masking}&\textbf{emotion modification strategy}&\textbf{sync $\uparrow$}&\textbf{visual quality $\uparrow$}& \textbf{$\Delta$ emotion $\uparrow$}\\
 \midrule
half&L1 + Emotion Objective&2.467&2.138&\textbf{0.779}\\
half&Emotion Objective&\textbf{3.492}&\textbf{2.887}&0.379\\
half&L1&2.758&1.85&0.732 \\
\midrule
\multicolumn{2}{c}{\textbf{emotion modification}}&\textbf{sync $\uparrow$}&\textbf{visual quality $\uparrow$}& \textbf{$\Delta$ emotion $\uparrow$}\\
\midrule
\multicolumn{2}{c}{happy $\rightarrow$ neutral}&2.483&2.117&\textbf{0.833}\\
\multicolumn{2}{c}{happy $\rightarrow$ sad}&2.35&1.967&0.569\\
\multicolumn{2}{c}{neutral $\rightarrow$ happy}&\textbf{3.483}&\textbf{2.933}&0.454\\
\multicolumn{2}{c}{neutral $\rightarrow$ sad}&2.95&2.183&0.533\\
\multicolumn{2}{c}{sad $\rightarrow$ happy}&3.333&2.167&0.621\\
\multicolumn{2}{c}{sad $\rightarrow$ neutral}&2.833&2.383&0.771\\

\bottomrule

\end{tabular}
\caption{User score from 1-5 over \textbf{Sync} and \textbf{Visual quality}, as well as \textbf{$\Delta$ emotion} of generated outputs normalized by expected change. Top half averaged over 3 best variants and bottom half averaged over emotions.}
\label{tab:user_study}
\end{table*}

We present metrics separately for each emotion modification pair of \textit{source} and \textit{destination emotions} in Table \ref{tab:per_emotion_results}. The top section is averaged over all model variants, while the bottom section presents the numbers for the best all-around performing model, \textbf{half} masking with \textbf{L1 + Emotion Objective} emotion modification strategy. We also report the baseline change in valence and arousal between ground truth inputs and targets ($\mathbf{v_d - v_s}$ and $\mathbf{a_d - a_s}$) to reveal the NISL2020 model's sensitivities to changes in each pair of emotions as performed by humans. 
Notably the difference between neutral and sad appears to manifest more as a change in arousal than as a change in valence. 
These baseline changes also demonstrates how the normalized $\Delta$ valence and $\Delta$ arousal are sensitive to noise in generated changes when there are relatively small changes of the ground truth ($\mathbf{v_d - v_s}$ and $\mathbf{a_d - a_s}$) in the denominator, possibly explaining erratic scores for arousal change between sad and happy and valence change between sad and neutral.

More generally Table \ref{tab:per_emotion_results} shows how emotion results are highly dependent on the source and destination emotion while synchronization and visual quality are relatively unaffected. Some \textit{destination emotions} may be easier to optimize for or some input \textit{source emotions} may be more difficult to alter. Notably the $\Delta ~\text{valence}$ scores for happy to sad are much higher than all others including sad to happy, possibly indicating that the model is able to exploit traits of sadness as it appears in the training data more effectively than other emotions. Nevertheless the best all around performing model gets $\Delta ~\text{valence}$  and $\Delta ~\text{arousal}$ values around $1.0$ on many emotion pairs, indicating a general ability to modify emotion comparably to the ground truth changes performed by the actors in the MEAD dataset.

\subsection{User Study}

We conduct a small crowdworker evaluation to rate the outputs of the top
3 models on their visual quality, lip synchronization, and emotion. We randomly select 2 videos per emotion pair for each of the 3 best performing models shown in the
top half of Table \ref{tab:top_level_results}, giving us 2 * 6 * 3 or 36 videos
overall.  We create a simple form, that asks 20 workers to rate each of these 36 videos on their visual quality and lip synchronization aspects, and to judge the emotion expressed by the speaker in the video. We select 20 workers from the Prolific platform \cite{palan2018prolific}, who are fluent in English, have participated in at least 10 tasks on Prolific, and have a 95\% acceptance score on earlier tasks.  For the first two factors, we ask workers to
rank the generated videos between  1 and 5, where 1 stands for "Very bad lip
synchronization" and "Very bad visual quality" and 5 stands for "Excellent lip
synchronization" and "Excellent visual quality." To judge the emotion expressed in the generated video, workers scored a video between 1 and 5
where a score of 1 meant that the speaker in the video is "Very Sad", a score of 3 meant that the speaker in the video is "Neutral" and a score of 5 meant that the speaker in the video is "Very Happy."

In order to make our emotion metric comparable across different pairs of \textit{source} and \textit{destination} emotions, we utilize a similar normalization scheme as on our automated emotion evaluations. The raw per-video emotion scores for generated outputs ($e_g$) are normalized with respect to the ground truth emotion values of the \textit{source} and \textit{destination} emotions ($e_s$ and $e_d$). So a modification from neutral to happy would have $e_s = 3$ and $e_d = 5$. We normalize the raw scores as follows:
\begin{align}
    \Delta ~\text{emotion} = {{e_g - e_s} \over {e_d - e_s}}
\end{align}
Thus positive values indicate change in the correct direction. And again, we further take $\Delta ~\text{emotion}_{*2n} = 1 - |1 - \Delta ~\text{emotion}|$ as our normalized score for neutral \textit{destination} modifications to penalize overshooting.

The user study results aggregated over model variants (top of Table \ref{tab:user_study}) corroborate findings of our novel automatic emotion evaluation by indicating that \textbf{L1} and \textbf{L1 + Emotion Objective} are most effective at emotion modification but struggle with synchronization and visual quality. The per-emotion pair results (bottom of Table \ref{tab:user_study}), show that we achieve meaningful emotion modification across emotion pairs along with moderate synchronization and visual quality.

\section{Conclusion}
To our knowledge, Wav2Lip-Emotion is the first approach to cast the synthesis of facial emotion as a video-to-video translation task. Our method extends an existing lip synchronization model, Wav2Lip, with a new task of modifying facial emotion in translated images through L1 reconstruction and pre-trained emotion objectives. In order compare results from all model variants on the MEAD dataset, we propose a novel automatic emotion evaluation approach and corroborate it with a user study.

Our evaluations support our ability to modify emotion, with both automatic metrics and human judgements rating the emotion modification in our best performing models as nearly comparable to the ground truth changes performed by the actors in the MEAD dataset. Our model also appears to maintain the level of lip synchronization present in the input videos, as was the original intent of building on the Wav2Lip architecture. However the visual quality of our models is only moderate. Greater performance may be possible from training on additional data with more variety of posses and speaker identities. While we have taken advantage of the paired emotion videos in the MEAD dataset for emotion evaluation, Wav2Lip-Emotion does not require such paired data. It can train on any videos in which a single speaker is found performing differing emotions. These could be labeled, as in the CMU-MOSEI dataset \cite{bagher-zadeh-etal-2018-multimodal}, or automatically inferred. Likewise while the present work has been limited to individual models for translating from a specific emotion to another one, our approach makes no fundamental obstacle to a multi-task approach that translates many emotions. By utilizing additional intensity labels, it may even be possible to dial the level of emotion modification through a hyperparameter, similarly to the approach of Fan et al. \cite{Fan_Huang_Gan_Huang_Gong_2019}. Such possibilities suggest the further promise of this proof of concept of video-to-video translation for modifying facial emotion.

\section{Acknowledgments}
This research was made possible by the generous support by the MIT Media Lab Consortium and the CISCO gift to Research in Telecreativity. We are grateful to Dr. William T. Freeman and Dr. Phillip Isola for their guidance and advice on this project. We also thank the anonymous ADGD 2021 reviewers for their insightful and helpful comments.

\bibliography{main.bib}

\appendix
\section{Hyperparameters}

All variants, except \textbf{Emotion Objective} without L1, use an L1 reconstruction loss weight ($s_r$) of 0.8, a synchronization loss weight ($s_w$) of 0.03, a visual quality discriminator weight ($s_g$) of 0.07, and an emotion objective weight ($s_e$) of 0.1. In the \textbf{Emotion Objective} without L1 variant, the emotion objective and L1 loss have contrary goals in that the target images whose reconstruction is measured by the L1 loss are from the \textit{source emotion} rather than the \textit{destination emotion} which is optimized for by the emotion objective. Thus in this variant we use an L1 reconstruction loss weight ($s_r$) of 0.6 and an emotion objective weight ($s_e$) of 0.3 to help the emotion modification compete with the L1 loss. These values were chosen based on the original hyperparameters of Wav2Lip, which demonstrates a need for a large L1 loss weight.

Meanwhile in the \textbf{L1} and \textbf{L1 + Emotion Objective} variants we utilize target videos from the \textit{destination emotion} while using videos from the \textit{source emotion} to provide the reference frames that inform the model of the identity of the speaker. However, while the MEAD data does contain emotion variations of the same utterances, the performances are not perfectly synchronized. Thus with the \textbf{full} masking models we also utilize the \textit{destination emotion} video for the masked pose prior input, as this allows the greatest pose synchronization. The \textbf{half} masking models however do not fully obscure the emotion information in the pose prior input and thus we use the \textit{source} instead of \textit{destination emotion} for the masked pose prior, despite the imperfect pose synchronization. 

We normalize inputs to the emotion objective by the channel mean and standard deviation of the MEAD data and convert to greyscale, as the emotion objective was pre-trained on greyscale only images. In order to further prevent vanishing gradient issues with the visual quality discriminator we clamp the gradient norm between 1e-2 and 1e10. All other settings are as specified by the original Wav2Lip. Hyperparameter optimization may increase performance but was forgone for this proof of concept due to limited compute and the long training times required. Experiments were run on a shared compute cluster running Ubuntu 20.04 utilizing GTX 1080 Ti, GTX TITAN X, and RTX 2080 GPUs, Intel Xeon CPU E5-2620 v4, Silver 4114, and Core i7-5930K CPUs, and RAM ranging from 64-768gb as availability permitted throughout the course of research. 

\section{Dataset}

We randomly separate the actors in the dataset into 37, 5, and 5 actor splits for training, validation, and testing. We also ensure that the validation and test splits contain at least 2 actors from the smaller of the two reported genders, and ensure that there is a spread in the representation of different ethnic backgrounds. In our test and validation set, we have 2 female and 3 male actors, while our training set has 27 male and 20 female actors.

The Wav2Lip architecture trains on windows of video frames and corresponding audio. Each frame in the window contains the face of the speaker alone. Videos of a speaker are broken down into their constituent frames, the faces are detected, cropped face frames are extracted, and stored along with the corresponding audio. As Wav2Lip is trained on 25 FPS LRS2 data we resample the MEAD data to 25 FPS as well.

\end{document}